\documentclass[11pt]{amsart}
\usepackage[margin=3cm]{geometry}
\usepackage{algorithm}
\usepackage{algorithmic}
\usepackage{graphicx}
\usepackage{booktabs}

\usepackage[utf8]{inputenc}
\usepackage{textcomp}

\usepackage[backend=bibtex, style=numeric]{biblatex}
\usepackage{csquotes}
\newcommand{\descrcell}[2]{
	\scriptsize
	\begin{tabular}[t]{@{}c@{}}\normalsize#1\\#2\end{tabular}
}

\begin{document}
\title[\resizebox{0.95\textwidth}{!}{Genetic and Memetic Algorithm with Diversity Equilibrium based on Greedy Diversification}]{Genetic and Memetic Algorithm with Diversity Equilibrium based on Greedy Diversification}

\author{Andrés Herrera-Poyatos $^1$ and Francisco Herrera $^{1,2}$}

\maketitle

\begin{center}
    $^1${\emph Research group ``Soft Computing and Intelligent Information Systems'', University of Granada, 18071 Granada, Spain.} \\
    $^2${\emph Department of Computer Science and Artificial Intelligence, University of Granada, 18071 Granada, Spain.} \\
    \emph{E-mail: {\tt andreshp9@gmail.com}, {\tt herrera@decsai.ugr.es}}
\end{center}

\begin{abstract}
    The lack of diversity in a genetic algorithm's population may lead to a bad performance of the genetic operators since there is not an equilibrium between exploration and exploitation. In those cases, genetic algorithms present a fast and unsuitable convergence.

	In this paper we develop a novel hybrid genetic algorithm which attempts to obtain a balance between exploration and exploitation. It confronts the diversity problem using the named greedy diversification operator. Furthermore, the proposed algorithm applies  a competition between parent and children so as to exploit the high quality visited solutions. These operators are complemented by a simple selection mechanism designed to preserve and take advantage of the population diversity.

	Additionally, we extend our proposal to the field of memetic algorithms, obtaining an improved model with outstanding results in practice.

	The experimental study shows the validity of the approach as well as how important is taking into account the exploration and exploitation concepts when designing an evolutionary algorithm. \\

\end{abstract}

\section{Introduction} \label{sec:intro}

	Optimization problems are a relevant topic of artificial intelligence. In order to solve these problems, computer scientists have found inspiration in nature, developing bio-inspired algorithms \cite{bio-inspired}, \cite{nature-inspired-mh} and, in particular, evolutionary algorithms \cite{evol-computation}.

	Genetic algorithms \cite{ga} are one of the most famous evolutionary algorithms. They are founded in the concepts of evolution and genetic. A solution to an optimization problem is view as a chromosome. Genetic algorithms maintain a population of chromosomes which evolves thanks to the selection, crossover and mutation operators. The evolution process ends when a predefined criteria is achieved.

	The equilibrium between exploration and exploitation is the key for success when designing an evolutionary algorithm. M. Crepinsek et al. \cite{exploration-exploitation} define exploration as ``the process of visiting entirely new regions of the search space'', whereas exploitation is ``the process of visiting those regions of the search space within the neighborhood of previously visited points''. If an heuristic is mainly focused in exploration, then it may not find the high quality neighbors of the promising visited solutions. Conversely, if an heuristic is mainly focused in exploitation, then it may not explore the regions of the search space which lead to most of the high quality solutions for the problem. Hence, our purpose is developing a genetic algorithm which intercalates the exploration and exploitation phases as needed, focusing the attention in the population diversity.

	The population diversity is one of the cornerstone of the genetic algorithms' performance. Note that a genetic algorithm's population converges if, and only if, the population diversity converges to zero. If this happens, then the heuristic has entered in a never-ending exploitation phase. We say that it has converged to a local optimum due to the lack capability for increasing the population diversity. Hence, the diversity problem -- maintaining a healthy population diversity -- is closely related to achieving a proper equilibrium between exploration and exploitation. There are various proposals of the specialized literature which address this problem \cite{diversity-adaptative-operators}.

	In this proposal we tackle the diversity problem formulating a diversification operator which introduces diversity to the population when it is needed. The inserted new chromosomes are generated by a randomized greedy algorithm. Afterwards, we use this operator to design an hybrid genetic algorithm, which is shown to maintain a stable population diversity. The hybridization between greedy randomized and genetic algorithms produces great results because the greedy chromosomes allow the heuristic to explore the promising regions of the search space. Hybridization of evolutionary algorithms with other heuristics is a common practice which helps to improve the evolutionary algorithms'  performance \cite{hybridization}, \cite{hybrid-ga-sa}. Furthermore, the proposed genetic algorithm use a competition between parent and children, similar to the one used by differential evolution \cite{de}, so as to exploit the high quality visited solutions. These operators are complemented by simple selection mechanism  which we call randomized adjacent selection and is designed to preserve and take advantage of the population diversity. We refer to the proposed algorithm as genetic algorithm with diversity equilibrium based on greedy diversification (GADEGD).

	In order to obtain an improved model, we also extend the previous algorithm to the field of memetic algorithms \cite{ma}. The new algorithm is called memetic algorithm with diversity equilibrium based on greedy diversification (MADEGD).

	We have developed an experimental study for each of both models using the traveling salesman problem \cite{tsp}, \cite{tsp-variations} as the case of study. In GADEGD's study we analyze its parameters and we match it against other state of the art genetic algorithms (CHC \cite{chc} and Micro-GA \cite{mga}) in terms of the solutions quality, the convergence to optimal solutions and the population diversity. Furthermore, we show how GADEGD's components contribute to its performance. In MADEGD's study we also analyze its parameters and compare it with GADEGD. Additionaly,  MADEGD is matched against other state of the art metaheuristics based on local search (GRASP \cite{grasp} and iterated greedy \cite{iterated-greedy}, \cite{ig-tsp}) from a triple perspective, the solutions quality, the population diversity and the number of calls to the local search.

	The remainder of this article is organized as follows. In Section \ref{sec:ga-ma}, we shortly introduce genetic and memetic algorithms. In Section \ref{sec:gadegd}, we study the diversity problem in genetic algorithms and we also present the greedy diversification operator, the other GADEGD's components and the corresponding experimental analysis. In Section \ref{sec:madegd}, we formulate MADEGD and show the associated experimental results. In Section \ref{sec:conclusion}, we point out the obtained conclusions.

\section{Genetic and memetic algorithms} \label{sec:ga-ma}

	In this section we briefly introduce genetic and memetic algorithms (Sections \ref{sec:ga-ma:ga} and \ref{sec:ga-ma:ma} respectively) and provide the pseudo-codes which are used in the experimental analysis. Lastly, we particularize in the application of these algorithms to the traveling salesman problem (Section \ref{sec:ga-ma:tsp}), which is employed as the case of study.

	\subsection{Genetic Algorithms} \label{sec:ga-ma:ga}

		Let $f$ be the objective function associated to an optimization problem, $f : S \rightarrow \mathbb{R}$, where $S$ is the set of all the possible solutions. The purpose is minimizing (resp. maximizing) $f$. Thus, a solution $s$ is better than another if its objective value $f(s)$ is smaller (resp. greater).

		Let $P_t$ be a finite subset of $S$. $P_t$ is called the population of the genetic algorithm. We can define a genetic algorithm as a population based metaheuristic \cite{mh-trends}, \cite{survey-mh}, \cite{metaheuristics} which uses the selection, crossover and mutation operators to obtain a new population $P_{t+1}$ from $P_t$. The process is repeated until a stopping criteria is achieved. Then, the best solution found or the best solution in the last population is returned.

		A genetic algorithm with the previous definition does not guarantee that there is a chromosome in the new population as good as the previous populations' chromosomes. However, this statement can be achieved applying the elitism criteria, appending the best solution in $P_t$, denoted $bs(P_t)$, to $P_{t+1}$. Afterwards, some models also delete the worst solution from $P_{t+1}$, denoted $ws(P_{t+1})$. Elitism has been proved to improve the genetic algorithm results in most cases, even theoretically \cite{ga:convergence}. Consequently, genetic algorithms with elitism are a popular model among computer scientists.

		\begin{algorithm}
			\begin{algorithmic}[1]
				\scriptsize
				\caption{BuildNewPopulation($P$)} \label{alg:ga-bp}
				\REQUIRE A population $P$.
				\STATE $P' \leftarrow \emptyset$
				\STATE Select $\frac{|P|}{2}$ pairs of chromosomes from $P$ using binary tournament selection	.
				\STATE Cross each pair, with a probability $p_c \in (0,1]$, getting two children if crossed.
				\STATE Add the new children and the pairs that have not been crossed to $P'$.
				\STATE Produce a mutation in each solution of $P'$ with a probability $p_m \in (0,1]$.
				\STATE Elitism: $ws(P') \leftarrow bs(P)$
				\RETURN $P'$
			\end{algorithmic}
		\end{algorithm}

		Algorithm \ref{alg:ga-bp} shows how a new population is built in a usual generational genetic algorithm with elitism. The binary tournament selection \cite{ga:selection} is a widely used selection scheme in genetic algorithms. The variables $p_c$ and $p_m$ are known as the crossover and mutation probability respectively. We have used the values $p_c = 0.7$ and $p_m = 0.1$ as it is common in the literature. From Algorithm \ref{alg:ga-bp} one can easily constructs a genetic algorithm, see Algorithm \ref{alg:ga}. However, this standard model may not work properly due to the lack of diversity in the population as it is shown in Section \ref{sec:gadegd}.

		\begin{algorithm}
			\begin{algorithmic}[1]
				\scriptsize
				\caption{Generational genetic algorithm with elitism} \label{alg:ga}
				\REQUIRE The population size, named $n$.
				\STATE Initialize $P_0$ with $n$ random elements from $S$.
				\STATE $t \leftarrow 0$
				\WHILE {stopping criteria is not achieved}
				\STATE $P_{t+1} \leftarrow BuildNewPopulation(P_t)$
				\STATE $t \leftarrow t+1$
				\ENDWHILE
				\RETURN $bs(P_t)$
			\end{algorithmic}
		\end{algorithm}

	\subsection{Memetic Algorithms} \label{sec:ga-ma:ma}

		Memetic algorithms hybridize evolutionary algorithms and local search procedures in order to obtain a model with a better exploration and exploitation. We will focus our attention in the subset of memetic algorithms in which the evolutionary scheme is carried out by a genetic algorithm.

		An usual hybridization consists in applying the local search once per each genetic algorithm iteration. The chromosome to which the local search is applied is the one with the best objective value among those population's solutions that have not been improved by the local search yet, what is indicated by a boolean variable. Other approaches apply the local search to each population element. However, these waste too much time improving low quality solutions. It is better to use the computational resources improving only the promising chromosomes as the first approach did.

		Memetic algorithms with high quality local searches usually outperform genetic algorithms. One of the reasons is that the local search improves the population quality introducing diversity at the same time. Hence, we could classify local search as an excellent mutation operator but with a high complexity cost. Furthermore, the evolutionary character of the algorithm implies that the local search is likely applied to better solutions as time passes, obtaining a good synergy.

		Algorithm \ref{alg:ma} shows a memetic algorithm's pseudo-code. It has two differences with Algorithm \ref{alg:ga}. First, the population is initialized with a randomized greedy algorithm, explained in Algorithm \ref{alg:gra}, so as to not apply the local search to random solutions. Otherwise, too much time would be consumed by the local search at the beginning of the algorithm. Secondly, the local search is applied once per iteration as we discussed before.

		\begin{algorithm}
			\begin{algorithmic}[1]
				\scriptsize
				\caption{Memetic algorithm} \label{alg:ma}
				\REQUIRE The population size, named $n$.
				\STATE Initialize $P_0$ with $n$ solutions obtained by a greedy randomized algorithm.
				\STATE $t \leftarrow 0$
				\WHILE {stopping criteria is not achieved}
				\STATE $P_{t+1} \leftarrow BuildNewPopulation(P_t)$
				\STATE Apply the local search to the best solution not previously improved of $P_{t+1}$ (if it exits).
				\STATE $t \leftarrow t+1$
				\ENDWHILE
				\RETURN $bs(P_t)$
			\end{algorithmic}
		\end{algorithm}

	\subsection{Application to the traveling salesman problem} \label{sec:ga-ma:tsp}

		We have used the traveling salesman problem as the case of study for our proposal. Given a complete and weighed graph, this problem consists in  obtaining the Hamiltonian cycle which minimize the sum of its edges' weighs. This sum is named the solution cost. Therefore, it is a minimization problem and the objective function provides the cost of each solution.

		We have chosen the traveling salesman problem because it is a classical NP Hard problem  which has been extensively employed to study heuristics in the specialized literature \cite{tsp-benchmarking}.

		Researchers have developed a huge amount of genetic operators for the traveling salesman problem \cite{ga:operators}. We use the well known crossover OX and exchange mutation which have shown a good performance in experimental studies.

		One of the best heuristics for the traveling salesman problem is a local search named Lin-Kernighan \cite{lk}. We have chosen a modern version \cite{lk-code} as the local search for the experimental study.

\section{GADEGD: Genetic algorithm with diversity equilibrium based on greedy diversification} \label{sec:gadegd}

	In this section we propose a novel genetic algorithm with the aim of obtaining a good balance between exploration and exploitation.

	First, we introduce a measure of the population diversity and we show the diversity problem in genetic algorithms. Secondly, we develop an operator to tackle the diversity 	problem, called the greedy diversification operator. Thirdly, we introduce the genetic algorithm with diversity equilibrium based on greedy diversification (GADEGD). At last, we show the experimental results of the proposal from a triple perspective: solutions quality, convergence to optimal solutions and population diversity.

	\subsection{Population diversity in genetic algorithms} \label{sec:gadegd:diversity}

		The diversity of a population is a measure of how different its chromosomes are. If the diversity is low, then the chromosomes are similar. On the other hand, if the diversity is high, then the chromosomes are quite different.

		We need a distance measure, $d : S \times S \rightarrow \mathbb{R}_0^+$, in order to quantify the differences between two solutions. Then, we can define the diversity of the population as the mean of the distance between all pairs of chromosomes, which can be written as following:
		\begin{gather*}
			D_t = \frac{\sum_{s,s' \in P_t} d(s,s')}{n(n-1)}
		\end{gather*}

		In the traveling salesman problem a good distance measure is the number of edges in which two chromosomes differ. The maximum distance between two chromosomes for this measure is the number of cities in the problem. Therefore, the same happens for the diversity measure proposed before.

\begin{figure}[H]
	        \centering
	        \includegraphics[width=10cm]{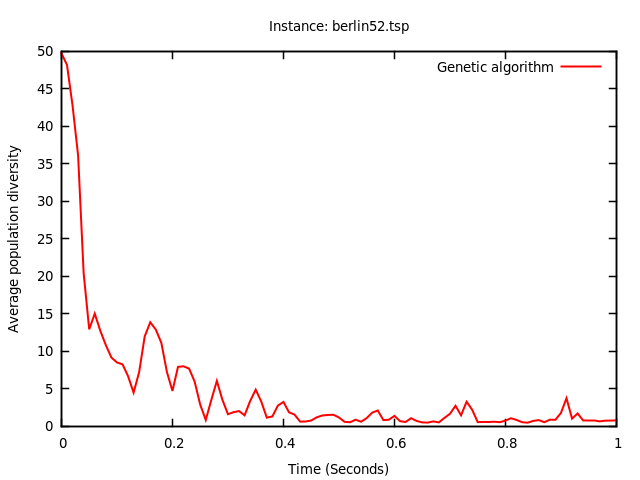}
            \caption{Diversity in a genetic algorithm's population (Algorithm \ref{alg:ga}).}
            \label{fig:diversity:ga}
        \end{figure}

		Figure \ref{fig:diversity:ga} shows how the population diversity evolves in a execution of a standard genetic algorithm (Algorithm \ref{alg:ga}). The instance is \textit{berlin52}, which consists of 52 cities and can be found in TSPLIB \cite{tsplib}. Each figure's point corresponds to the average population diversity in the last $0.01$ seconds. The diversity starts near the maximum possible value since the initial chromosomes are randomly chosen. Afterwards, the diversity quickly decreases because the algorithm focuses the search in a specific region of the search space. However, the diversity diminution is excessive, converging to a number close to zero eventually. This fact indicates that the algorithm has converged to a local optimum, not being able to reach better solutions. Consequently, if the local optimum is not good enough, then the algorithm results will be disappointing. We aim to avoid this fast and unsuitable convergence so as to improve the algorithm performance.

		In genetic algorithms, the population diversity is maintained by the mutation operator. The diversity depends on the value $p_m$ which was defined as the probability of mutating a chromosome in an iteration. If $p_m$ is equal to zero, then the diversity will tend to zero after few iterations. If $p_m$ is increased, then the diversity will converge to a higher value. Nonetheless, the mutation operator introduces diversity at the cost of deteriorating, most of the time, the quality of the solutions to which it is applied. Hence, low values are assigned to the mutation probability in the specific literature (between $0.1$ and $0.2$ per chromosome) not allowing a high diversity as it is shown in Figure \ref{fig:diversity:ga} (where $p_m$ is $0.1$).

	\subsection{Greedy diversification operator} \label{sec:gadegd:gd}

		Population diversity is a double-edged sword. It is needed to explore the solutions space but it can imply not finishing the exploration process. If it is the case, then not enough time is dedicated to the exploitation phase which is essential to get higher quality solutions. Therefore, it is desired a diversification operator that only introduces diversity if it is necessary.

		This operator would be applied to every new population as it is shown in Algorithm \ref{alg:gad}.

		\begin{algorithm}
			\begin{algorithmic}[1]
				\scriptsize
				\caption{Genetic algorithm with diversification} \label{alg:gad}
				\REQUIRE The population size, named $n$.
				\STATE Initialize $P_0$ with $n$ random elements from $S$.
				\STATE $t \leftarrow 0$
				\WHILE {stopping criteria is not achieved}
				\STATE $P_{t+1} \leftarrow BuildNewPopulation(P_t)$
				\STATE $P_{t+1} \leftarrow Diversification(P_{t+1})$
				\STATE $t \leftarrow t+1$
				\ENDWHILE
				\RETURN $bs(P_t)$
			\end{algorithmic}
		\end{algorithm}

		The diversification operator should delete the population's repeated chromosomes because they waste the population's slots and reduce the diversity. Furthermore, the chromosomes that are left in the population should have a good objective value and be potentially good for the crossover operator. The diversification operator ought also to have a low computational cost since the optimization is done by the evolutionary scheme. We propose using a greedy randomized algorithm to obtain chromosomes satisfying these conditions.

		Greedy randomized algorithms provide acceptable chromosomes from the objective value perspective that also contain high quality genetic material thanks to the greedy selection function. The randomized aspect of the algorithm supplies the diversity required in the generated solutions. There are some conditions to implement a greedy randomized algorithm for an optimization problem. First, the solution must be represented as a set or list of elements. Secondly, it is needed a greedy function which provides the quality of an element according to those that  have been already added to the solution. The building process is iterative. In each step a new element is added to the solution until it is fully completed. In order to add a new element, a restricted candidate list (RCL) must be determined. Afterwards, an element randomly chosen from the RCL is added to the solution. This process is presented in Algorithm \ref{alg:gra}.

		\begin{algorithm}
			\begin{algorithmic}[1]
				\scriptsize
				\caption{Greedy Randomized Algorithm} \label{alg:gra}
				\STATE $solution \leftarrow \emptyset$
				\WHILE {solution is not finished}
				\STATE Build the $RCL$.
				\STATE $x \leftarrow randomElement(RCL)$
				\STATE $solution \leftarrow solution \cup x$
				\STATE Adapt the greedy function to the new partial solution.
				\ENDWHILE
				\RETURN $solution$
			\end{algorithmic}
		\end{algorithm}

		The RCL contains the best elements conforming to the greedy function. The list's size can be constant or variable, in which case it depends on the elements quality. The variable size RCL contains the elements whose greedy value is less than $(1+\sigma)$ times the best element's value, where $\sigma$ is a fixed real value greater than zero. This model obtains better solutions because it controls the quality of the elements added to the list. It also keeps the diversity in the generated solutions since the RCL can be very large when multiple elements are good enough. In our experiments we use $\sigma = 0.1$ although this parameter can be optimized in each application domain.

		Particularizing in the traveling salesman problem, a solution is conceived as a list of nodes. The greedy function provides the distance of each node which is not in the solution to the last node appended to the solution. Thus, a node is better than another one if its distance to the last node appended is smaller. As a consequence, the obtained solution is mostly comprised of short edges. Therefore, if we cross this greedy solution with another one, we get a child which has a fair number of short edges and, hence, it is probably a high quality solution.

		Thus, the first element of our proposal is a diversification operator which uses the greedy randomized algorithm to substitute those chromosomes that share similarity characteristics with other population solutions. This procedure increase the diversity and also keeps the population quality. In order to formalize the operator, let us consider an arbitrary characteristic featured in the problem's solutions and let $C$ be the set of all its possible values. The function $g: S \rightarrow C$ provides, given a solution $s$, the value $g(s) \in C$ which the solution possesses. For instance, a characteristic could be the solution's objective value or whether the solution has a concrete element or not. It could even be the solution itself.

		\begin{algorithm}
			\begin{algorithmic}[1]
				\scriptsize
				\caption{Greedy diversification operator} \label{alg:gd}
				\REQUIRE The genetic algorithm population, named $P$, and the characteristic function $g: S \rightarrow C$.
				\STATE $P' \leftarrow \emptyset$
				\STATE Sort $P$ by the objective function (the better solutions are placed first).
				\STATE $k \leftarrow 0$
				\FOR{$s \textbf{ in } P$}
				\IF{exists $s'$ in $P'$ such as $g(s) = g(s')$}
				\STATE $k \leftarrow k + 1$
				\ELSE
				\STATE $P' \leftarrow P' \cup s$
				\ENDIF
				\ENDFOR
				\FOR{$i=1$ \TO $k$}
				\STATE $P' \leftarrow P' \cup GreedyRandomizedAlgorithm()$
				\ENDFOR
				\RETURN $P'$
			\end{algorithmic}
		\end{algorithm}

		Algorithm \ref{alg:gd} uses this terminology to show a general definition of the greedy diversification operator. This operator removes the population's worst solutions that share the characteristic's value with other ones. Then, it fills the new population with greedy randomized solutions. The efficiency in the worst case is $\theta(\vert P \vert (\log \vert P \vert + \phi + \mu))$, where $\phi$ and $\mu$ are the complexity of applying $g$ to a solution and obtaining a greedy randomized solution respectively.

		The choice of $g$ affects the amount of diversity introduced and the operator complexity. A first approach is using the identity function ($Id: S \rightarrow S$ and $Id(s)=s$) as $g$. In this case the algorithm just substitutes the repeated solutions in the population. Algorithm \ref{alg:gd:id} provides an efficient implementation for this approach. In the case of the traveling salesman problem, we can implement the identity function and the greedy randomized algorithm with efficiencies $\theta(m)$ and $\theta(m^2)$ respectively, where $m$ is the number of nodes in the instance. Consequently, the efficiency in the worst case is $\theta(\vert P \vert (\log \vert P \vert + m^2))$. However, the experimental analysis in Section \ref{sec:gadegd:experiments} shows that Algorithm \ref{alg:gd:id} complexity in practice is $O(\vert P \vert \log \vert P \vert + m^2)$ since two solutions usually have different objective values and few repeated solutions are found after a genetic algorithm's iteration.

		\begin{algorithm}
			\begin{algorithmic}[1]
				\scriptsize
				\caption{Greedy diversification operator with $g = Id$} \label{alg:gd:id}
				\REQUIRE The genetic algorithm population, named $P$.
				\STATE $P' \leftarrow \{bs(P)\}$
				\STATE Sort $P$ by the objective function (the better solutions are placed first).
				\FOR{$i = 1$ \TO $n-1$ }
				\IF{$f(P[i-1]) = f(P[i])$ and $P[i-1] = P[i]$}
				\STATE $P' \leftarrow P' \cup GreedyRandomizedAlgorithm()$
				\ELSE
				\STATE $P' \leftarrow P' \cup P[i]$
				\ENDIF
				\ENDFOR
				\RETURN $P'$
			\end{algorithmic}
		\end{algorithm}

		A second approach is using the objective function as $g$. In this case more diversity is introduced but some interesting solutions might be lost. The implementation is the same that the one given in Algorithm \ref{alg:gd:id} but without comparing two solutions in the line 4. The practical complexity remains the same too. Both approaches' results are contrasted in Section \ref{sec:gadegd:experiments:parameters}.

	\subsection{Genetic algorithm with diversity equilibrium based on greedy diversification} \label{sec:gadegd:gadegd}

		Algorithm \ref{alg:gad} with the greedy diversification operator given in Algorithm \ref{alg:gd:id} presents a much better performance than Algorithm \ref{alg:ga} as we show in Section \ref{sec:gadegd:experiments}. However, the synergy among the genetic  and diversification operators can be improved. Therefore, we propose a novel genetic algorithm with the following characteristics:

		\begin{enumerate}
			\item A novel selection mechanism which does not apply pressure and helps to preserve the diversity in the new population. We call it randomized adjacent selection.
			\item The crossover probability is equal to 1.
			\item A competition between parent and children to increase the pressure applied to the population.
			\item The greedy diversification operator is used instead of the mutation operator.
		\end{enumerate}

		The new algorithm is named genetic algorithm with diversity equilibrium based on greedy diversification since it gets a healthy diversity thanks to the greedy diversification operator and it is referred as GADEGD. The mentioned algorithm's components are explained in the rest of the section.

		Selection schemes in genetic algorithms usually ignore the population's worst solutions. Some examples are the tournament or ranking selection \cite{ga:selection}, which select the worst solutions with a very low probability. If we use these mechanisms, then the greedy solutions introduced by the diversification operator will not be selected eventually. Furthermore, we desire every chromosome to be crossed in order to take advantage of the population diversity. As a consequence, we propose randomly sorting the population and crossing the adjacent solutions, considering the first and last solution also as contiguous. Each pair of adjacent solutions is crossed with probability 1, generating only one child. We call it randomized adjacent selection. Note that this scheme assures that each solution has exactly two children. Consequently, all the genetic material is used to build the new population, what preserves the diversity.

		The randomized adjacent selection conserves the diversity but does not apply any pressure to the population. The competition between parent and children is the mechanism chosen for that purpose. We propose a process similar to the one used by the differential evolution algorithms; each child only competes with its left parent and the best of both solution is added to $P_{t+1}$. Consequently, the population $P_{t+1}$ contains a descendant for each solution of $P_t$ or the solution itself. This statement implies that if the population $P_t$ is diverse, then the population $P_{t+1}$ will likely be diverse too. Furthermore, the population $P_{t+1}$ is always better than $P_t$ in terms of the objective function. The competition between parent and children can be considered a strong elitism that, in our case, preserves the diversity thanks to the randomized adjacent selection.

		\begin{algorithm}
			\begin{algorithmic}[1]
				\scriptsize
				\caption{BuildNewPopulationGADEGD($P$)} \label{alg:gadegd-bp}
				\REQUIRE A population $P$.
				\STATE $P' \leftarrow \emptyset$
				\STATE Sort $P$ randomly.
				\FOR{$i = 0$ \TO $n-1$}
				\STATE $parent1 \leftarrow P(i)$
				\STATE $parent2 \leftarrow P( (i+1) \mod n )$
				\STATE $child \leftarrow Crossover(parent1,parent2)$
				\IF{$f(child) \text{ is better than } f(parent1)$}
				\STATE $P' \leftarrow P' \cup child$
				\ELSE
				\STATE $P' \leftarrow P' \cup parent1$
				\ENDIF
				\ENDFOR
				\RETURN $P'$
			\end{algorithmic}
		\end{algorithm}

		Algorithm \ref{alg:gadegd-bp} shows how a new population is built in GADEGD. Note that the code is very simple, what is an advantage versus more complicated models.

		In genetic algorithms, the mutation operator introduces diversity and allows the algorithm to explore the neighborhood of the population's solutions. However, GADEGD does not need it any more since it is able to keep the population diversity by itself. Consequently, the mutation operator just decrease the solutions quality and should not be used. Algorithm \ref{alg:gadegd} contains the pseudo-code of GADEGD.

		\begin{algorithm}
			\begin{algorithmic}[1]
				\scriptsize
				\caption{Genetic algorithm with diversity equilibrium based on greedy diversification} \label{alg:gadegd}
				\REQUIRE The population size, named $n$, and the characteristic function $g: S \rightarrow C$.
				\STATE Initialize $P_0$ with $n$ random elements from $S$.
				\STATE $t \leftarrow 0$
				\WHILE {stopping criteria is not achieved}
				\STATE $P_{t+1} \leftarrow BuildNewPopulationGADEGD(P_t)$
				\STATE $P_{t+1} \leftarrow GreedyDiversification(P_{t+1}, g)$
				\STATE $t \leftarrow t+1$
				\ENDWHILE
				\RETURN $bs(P_t)$
			\end{algorithmic}
		\end{algorithm}

\begin{figure}[H]
        	\centering
        	\includegraphics[width=10cm]{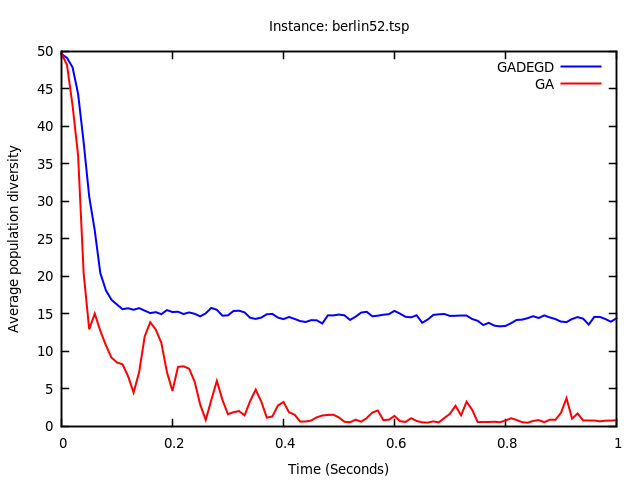}
        	\caption{Diversity: GADEGD vs generational genetic algorithm (Algorithm \ref{alg:ga})}
        	\label{fig:diversity:gadegd}
        \end{figure}

		Figure \ref{fig:diversity:gadegd} shows how the population diversity evolves for GADEGD and the implemented genetic algorithms (Algorithms \ref{alg:gadegd} and \ref{alg:ga} respectively) in the instance \textit{berlin52}. Here GADEGD has been executed with $g = Id$. Note that GADEGD is designed to maintain a diverse population and so it does. The initial diversity decreases quickly in both algorithms. Afterwards, GADEGD keeps the diversity in a high and stabilized value. Its components allows the algorithm to work with good solutions in multiple zones of the search space. Besides, if the population diversity decreases, then the greedy diversification introduces new chromosomes.

	\subsection{Experimental analysis} \label{sec:gadegd:experiments}

		The experiments were done in a computer with 8 GB of RAM and a processor Intel I5 with 2.5 GHz. The 18 instances of the traveling salesman problem can be found in the TSPLIB library. Each result is computed as the average of 30 executions.

		The experimental analysis contains 3 subsections. First, we provide a study of the GADEGD's parameters: the population size and the characteristic function. In the second subsection the algorithm is compared against other state of the art algorithms from a triple perspective: the solutions quality, the convergence to the instances' optimums and the population diversity. Lastly, we analyze how much the GADEGD's components contribute to its performance.

		\subsubsection{GADEGD's parameters analysis} \label{sec:gadegd:experiments:parameters}

			The population size has a huge impact on a genetic algorithm behavior. On the one hand, a greater population size contribute to the exploration of the solutions' space, avoiding a fast and unsuitable convergence. However, a large population needs much more computational time to exploit the most promising solutions. On the other hand, a smaller population size implies a higher exploitation and a sooner convergence. The optimal population size depends on the execution's time and the algorithm facilities to maintain a diverse population. If this optimal value is very large, then the algorithm has probably difficulties to explore the solutions space and keep the population diversity. If this is the case, then the algorithm is probably improvable.

			Genetic algorithms are usually assigned a population size between $30$ and $100$ in the literature although this value tends to grow with the improvements in hardware. There are also models which work under small populations \cite{mga}. In our case, we want the algorithm to have a medium sized population because we try to achieve an equilibrium between exploration and exploitation.

			Table \ref{table:gadegd:tp} compares the population sizes $32$ and $64$ in terms of the mean and standard deviation of the obtained solutions' objective value. In these experiments the GADEGD's characteristic function is $g=Id$ and the execution's time is $0.1 m$ seconds, where $m$ is the instance's number of nodes. The experiments show that $64$ is a better population size than $32$, obtaining the best results most of the time. We also have executed the algorithm with smaller and larger population sizes and they had a significant worse performance. Consequently, we are using $64$ as the standard population size for the GADEGD algorithm.

			\begin{table}[ht]
             	\caption{GADEGD with $g = Id$ and population sizes $32$ and $64$. The execution time is $0.1 m$ seconds. The best results are highlighted in bold. The last row indicates the number of times that each model got the best result in any instance.}
             	\label{table:gadegd:tp}
             	\centering
             	\kern 2mm
             	\begin{tabular}{llllll}
             		\toprule
             		\textbf{Problem} & \textbf{Optimum} & \multicolumn{2}{c}{\textbf{Mean objective value}} & \multicolumn{2}{c}{\textbf{Standard deviation}}\\
             		\midrule
             		\multicolumn{2}{c}{} & $n = 32$ & $n = 64$ & $n = 32$ & $n = 64$ \\
             		\cmidrule{3-6}
             		eil51    & 426    & \textbf{427.2} & 427.267 & 0.4 & 1.26315 \\
             		berlin52 & 7542   & \textbf{7555.1} & 7572.57 & 39.3 & 55.4068 \\
             		st70     & 675    & 682.467 & \textbf{682.067} & 5.03808 & 4.21057 \\
             		eil76    & 538    & 549.8 & \textbf{549.5} & 0.979796 & 1.43178 \\
             		pr76     & 108159 & \textbf{109169} & 109395 & 942.768 & 573.954 \\
             		kroA100  & 21282  & 21385.8 & \textbf{21352.5} & 134.201 & 119.631 \\
             		rd100    & 7910   & \textbf{7917.13} & 7919.47 & 28.619 & 35.4445 \\
             		eil101   & 629    & \textbf{631.7} & 633.3 & \textbf{3.06757} & 4.09186 \\
             		lin105   & 14379  & 14436.2 & \textbf{14430.5} & 33.7544 & 18.2916 \\
             		ch150    & 6528   & 6588.67 & \textbf{6578.67} & 20.277 & 20.6968 \\
             		rat195   & 2323   & 2393.47 & \textbf{2386.83} & 16.126 & 17.6335 \\
             		d198     & 15780  & 16076.4 & \textbf{16053.9} & 75.0803 & 95.1888 \\
             		ts225    & 126643 & 127550 & \textbf{127427} & 391.036 & 524.795 \\
             		a280     & 2579   & 2718.63 & \textbf{2704.5} & 26.2824 & 26.9923 \\
             		lin318   & 42029  & 43815.3 & \textbf{43739.5} & 429.102 & 412.694 \\
             		fl417    & 11861  & 12323.7 & \textbf{12303.9} & 105.208 & 134.313 \\
             		pcb442   & 50778  & 55741 & \textbf{55502} & 633.782 & 639.032 \\
             		rat575   & 6773   & 7677.5 & \textbf{7670.97} & 83.7412 & 78.0105 \\
             		\midrule
             		& & 5  & 13 \\
             		\cmidrule[0.3mm]{3-6}
             	\end{tabular}
            \end{table}

            \begin{table}[ht]
            	\caption{GADEGD with population size $64$ and the characteristic functions $g=Id$ and $g=f$. The execution time is $0.1 m$ seconds. The best results are highlighted in bold. The last row indicates the number of times that each model got the best result in any instance.}
            	\label{table:gadegd:function}
            	\centering
            	\kern 2mm
            	\begin{tabular}{llllll}
            		\toprule
            		\textbf{Problem} & \textbf{Optimum} & \multicolumn{2}{c}{\textbf{Mean objective}} & \multicolumn{2}{c}{\textbf{Percent of explored}}\\
            		 & & \multicolumn{2}{c}{\textbf{value}}  & \multicolumn{2}{c}{\textbf{solutions generated}}\\
            		 & & \multicolumn{2}{c}{} & \multicolumn{2}{c}{\textbf{in the greedy}}\\
            		 & & \multicolumn{2}{c}{} & \multicolumn{2}{c}{\textbf{diversification}}\\
            		\midrule
            		\multicolumn{2}{c}{} & $g = Id$ & $g = f$ & $g = Id$ & $g = f$ \\
            		\cmidrule{3-6}
            		eil51    & 426    & \textbf{427.267} & 428.4 & 5.36748 & 10.1318 \\
            		berlin52 & 7542   & \textbf{7572.57} & \textbf{7572.57} & 5.08438 & 5.23617 \\
            		st70     & 675    & \textbf{682.067} & 686.433 & 4.62881 & 7.36705 \\
            		eil76    & 538    & \textbf{549.5} & 549.9 & 4.27079 & 8.90469 \\
            		pr76     & 108159 & \textbf{109395} & 109503 & 4.55236 & 4.57542 \\
            		kroA100  & 21282  & \textbf{21352.5} & 21384.6 & 4.90119 & 4.91509 \\
            		rd100    & 7910   & \textbf{7919.47} & 7926.5 & 4.56272 & 4.70796 \\
            		eil101   & 629    & \textbf{633.3} & 634.233 & 5.28149 & 7.87105 \\
            		lin105   & 14379  & 14430.5 & \textbf{14423.3} & 4.82928 & 4.8212 \\
            		ch150    & 6528   & \textbf{6578.67} & 6586.03 & 4.30821 & 4.49253 \\
            		rat195   & 2323   & \textbf{2386.83} & 2400.87 & 3.88133 & 4.52266 \\
            		d198     & 15780  & \textbf{16053.9} & 16076.2 & 4.18145 & 4.261 \\
            		ts225    & 126643 & 127427 & \textbf{127355} & 2.7799 & 2.59045 \\
            		a280     & 2579   & \textbf{2704.5} & 2727.53 & 4.09656 & 4.68289 \\
            		lin318   & 42029  & \textbf{43739.5} & 43843.7 & 3.97038 & 4.03494 \\
            		fl417    & 11861  & 12303.9 & \textbf{12286} & 4.01814 & 4.09853 \\
            		pcb442   & 50778  & 55502 & \textbf{55477} & 2.72923 & 2.74129 \\
            		rat575   & 6773   & \textbf{7670.97} & 7712.5 & 2.08298 & 2.21214 \\
             		\midrule
             		& & 14 & 5 \\
             		\cmidrule[0.3mm]{3-6}
            	\end{tabular}
            \end{table}

			The most essential GADEGD's parameter is the characteristic function. We have used the functions $g = Id$ and $g = f$ explained in Section \ref{sec:gadegd:gd}. More complex models did not obtained better results in practice. Table \ref{table:gadegd:function} compares both functions' performance. The model $g=Id$ reaches better solutions in most instances. The function $g=f$ introduces too much diversity and it might substitute not repeated chromosomes with unique characteristics. Hence, the model $g=Id$ is the one chosen for the rest of the study.

			Table \ref{table:gadegd:function} also shows the percent of explored solutions which are generated in the greedy diversification. This value is usually between 2 and 10 \%. In average, this means that the algorithm introduces between 1 and 7 greedy solutions per iteration for both characteristics functions. Consequently, we can consider the practical complexity of these greedy diversification algorithms as $O(\vert P \vert \log \vert P \vert + m^2)$ as we mentioned in Section \ref{sec:gadegd:gd}. Note that if the GADEGD algorithm converges, then the greedy diversification introduces more greedy solutions to increase the population diversity. If it is not the case, then less greedy solutions are introduced (see instances 1 and 18 respectively in Table \ref{table:gadegd:function}).

		\subsubsection{Comparison with other genetic algorithms which use diversity mechanisms instead of mutations} \label{sec:gadegd:experiments:state-of-the-art}

			In this section we compare GADEGD with the genetic algorithm given in Algorithm \ref{alg:ga} and other recognized models which do not use the mutation operator: CHC \cite{chc} and Micro-GA \cite{mga}. We study the quality of the obtained solutions, the convergence to the problems' optimums and the population diversity in order to illustrate GADEGD's performance.

			CHC was the first genetic algorithm which applies a competition between parent and children. CHC has already been applied to the traveling salesman problem variations \cite{chc:tsp}. Our implementation has the following characteristics:

			\begin{itemize}
				\item Population size = 60
				\item Random selection with incest prevention mechanism that avoids crossing similar solutions.
				\item Competition between parent and children: the population $P_{t+1}$ contains the best chromosomes between parent and children.
				\item Reinitialization of the population when it converges (detected by the incest prevention mechanism): the best chromosome is left and the other ones are replaced by random solutions.
			\end{itemize}

			The Micro-GA was proposed as a genetic algorithm with a small population and fast convergence. It was the first genetic algorithm which uses a reinitialization of the population when it converges. It has the following characteristics:

			\begin{itemize}
				\item Population size = 5
				\item The best solution in $P_t$ is added to $P_{t+1}$.
				\item Two pairs of parent are selected by a variation of the tournament selection.
				\item Both pairs are crossed, generating two children per pair that are added to $P_{t+1}$.
				\item Reinitialization of the population when it converges (all the solutions have the same objective value): the best chromosome is left and the other ones are replaced by random solutions.
			\end{itemize}

            \begin{table}[H]
            	\caption{CHC and Migro-GA compared against the same models with greedy reinitialization. The execution time is $0.1 m$ seconds. The best results are highlighted in bold. The last row indicates the number of times that each model got a better result than the same algorithm with a different reinitialization.}
            	\label{table:chc-mga}
            	\centering
            	\kern 2mm
            	\begin{tabular}{llllll}
            		\toprule
            		\textbf{Problem} & \textbf{Optimum} & \multicolumn{4}{c}{\textbf{Mean objective value}} \\
            		\midrule
            		\multicolumn{2}{c}{} & \multicolumn{2}{c}{CHC} & \multicolumn{2}{c}{Micro-GA} \\
            		\cmidrule{3-6}
            		\multicolumn{2}{c}{} & \multicolumn{1}{c}{Classical} & \multicolumn{1}{c}{Greedy} & \multicolumn{1}{c}{Classical} & \multicolumn{1}{c}{Greedy} \\
            		\multicolumn{2}{c}{} & \multicolumn{1}{c}{model} & \multicolumn{1}{c}{Reinitialization} & \multicolumn{1}{c}{model} & \multicolumn{1}{c}{Reinitialization} \\
            		\cmidrule{3-6}
            		eil51    & 426    & 496.8 & \textbf{443.933} & 447.267 & \textbf{432.267} \\
            		berlin52 & 7542   & 8041.8 & \textbf{7633.1} & 8053.27 & \textbf{7588.1} \\
            		st70     & 675    & 889 & \textbf{730.8} & 743.967 & \textbf{689.9} \\
            		eil76    & 538    & 665.2 & \textbf{574} & 586.933 & \textbf{554.767} \\
            		pr76     & 108159 & 114084 & \textbf{109572} & 116626 & \textbf{111017} \\
            		kroA100  & 21282  & 24010.4 & \textbf{21377.7} & 25627.2 & \textbf{21689.2} \\
            		rd100    & 7910   & 9496.23 & \textbf{7998.23} & 9418.3 & \textbf{8010.03} \\
            		eil101   & 629    & 827.7 & \textbf{686.267} & 729.433 & \textbf{634.567} \\
            		lin105   & 14379  & 19445.4 & \textbf{14426.4} & 16827.7 & \textbf{14511.4} \\
            		ch150    & 6528   & 9311.93 & \textbf{6763.63} & 9294.63 & \textbf{6626.73} \\
            		rat195   & 2323   & 3515.8 & \textbf{2431.4} & 3566.53 & \textbf{2430.53} \\
            		d198     & 15780  & 21395.6 & \textbf{16603.7} & 22493.2 & \textbf{16387.3} \\
            		ts225    & 126643 & 214322 & \textbf{133175} & 251757 & \textbf{129840} \\
            		a280     & 2579   & 5109.53 & \textbf{2891.43} & 5496.5 & \textbf{2791.93} \\
            		lin318   & 42029  & 82239.7 & \textbf{43917.3} & 107959 & \textbf{44886.4} \\
            		fl417    & 11861  & 32020.3 & \textbf{12937} & 60670.7 & \textbf{12674.2} \\
            		pcb442   & 50778  & 117600 & \textbf{57568.3} & 174361 & \textbf{58950.6} \\
            		rat575   & 6773   & 18170.7 & \textbf{7773.43} & 26568 & \textbf{8024.4} \\
             		\midrule
             		& & 0 & 18 & 0 & 18 \\
             		\cmidrule[0.3mm]{3-6}
            	\end{tabular}
            \end{table}

            Both algorithms assign $1$ to the crossover probability and do not use the mutation operator. In this sense, they are similar to our proposal. However, they use a reinitialization of the population in contrast to GADEGD's greedy diversification.

			Table \ref{table:chc-mga} shows the results obtained by these algorithms. They are good in instances with few nodes. However, if the instances are harder, then they do not perform well, the random solutions are not good enough as a reinitialization mechanism. Consequently, we propose a greedy reinitialization for CHC and Micro-GA, replacing the population by greedy solutions obtained from Algorithm \ref{alg:gra} instead of random chromosomes. The results are also presented in Table \ref{table:chc-mga}. As we expected, the new models with the greedy reinitialization outperform the older ones in any instance. This fact shows that genetic algorithms hybridize fairly well with greedy algorithms, there is a great synergy between the greedy chromosomes and the crossover operator as we mentioned in Section \ref{sec:gadegd:gd}.

            \begin{table}[ht]
            	\caption{Comparison among GADEGD, GA, CHC and Micro-GA with greedy reinitialization in terms of the solutions' quality. The execution time is $0.1 m$ seconds. The best results are highlighted in bold and the worst are underlined. The last row indicates the number of times that each model got the best and worst result in any instance.}
            	\label{table:state-of-the-art:objective-value}
            	\centering
            	\kern 2mm
            	\begin{tabular}{llllll}
            		\toprule
            		\textbf{Problem} & \textbf{Optimum} & \multicolumn{4}{c}{\textbf{Mean objective value}} \\
            		\midrule
            		\multicolumn{2}{c}{} & GADEGD & GA & \descrcell{CHC}{with G.R.} & \descrcell{Micro-GA}{with G.R.} \\
            		\cmidrule{3-6}
            		eil51    & 426    & \textbf{427.267} & \underline{507.7} & 443.933 & 432.267 \\
            		berlin52 & 7542   & \textbf{7572.57} & \underline{9146.5} & 7633.1 & 7588.1 \\
            		st70     & 675    & \textbf{682.067} & \underline{910.067} & 730.8 & 689.9 \\
            		eil76    & 538    & \textbf{549.5} & \underline{692.4} & 574 & 554.767 \\
            		pr76     & 108159 & \textbf{109395} & \underline{146800} & 109572 & 111017 \\
            		kroA100  & 21282  & \textbf{21352.5} & \underline{34162.4} & 21377.7 & 21689.2 \\
            		rd100    & 7910   & \textbf{7919.47} & \underline{12421.3} & 7998.23 & 8010.03 \\
            		eil101   & 629    & \textbf{633.3} & \underline{886.833} & 686.267 & 634.567 \\
            		lin105   & 14379  & 14430.5 & \underline{24862.4} & \textbf{14426.4} & 14511.4 \\
            		ch150    & 6528   & \textbf{6578.67} & \underline{12898.4} & 6763.63 & 6626.73 \\
            		rat195   & 2323   & \textbf{2386.83} & \underline{5246.8} & 2431.4 & 2430.53 \\
            		d198     & 15780  & \textbf{16053.9} & \underline{35969.5} & 16603.7 & 16387.3 \\
            		ts225    & 126643 & \textbf{127427} & \underline{394841} & 133175 & 129840 \\
            		a280     & 2579   & \textbf{2704.5} & \underline{8280.07} & 2891.43 & 2791.93 \\
            		lin318   & 42029  & \textbf{43739.5} & \underline{154852} & 43917.3 & 44886.4 \\
            		fl417    & 11861  & \textbf{12303.9} & \underline{97497.8} & 12937 & 12674.2 \\
            		pcb442   & 50778  & \textbf{55502} & \underline{233595} & 57568.3 & 58950.6 \\
            		rat575   & 6773   & \textbf{7670.97} & \underline{35167.7} & 7773.43 & 8024.4 \\
            		\midrule
            		& & 17 / 0 & 0 / 18 & 1 / 0 & 0 / 0\\
            		\cmidrule[0.3mm]{3-6}
            	\end{tabular}
            \end{table}

			Table \ref{table:state-of-the-art:objective-value} compares, in terms of the solution's quality, the algorithms GADEGD, a generational genetic algorithm (Algorithm \ref{alg:ga}) and both CHC and Micro-GA with greedy reinitialization. GADEGD performs considerably better than the other algorithms in $17$ out of $18$ instances. The main reason behind the better performance of GADEGD is the greedy diversification. It introduces diversity before the algorithm has totally converged and, consequently, it constantly keeps a high quality and diverse population, what can not been achieved by the (greedy) reinitialization used in CHC and Micro-GA. In those algorithms the diversity and the quality of the solutions are not stable and they generally vary inversely until the population is reinitialized. When the population is reinitialized, the algorithms do a lot of effort to build a high quality population again and, consequently, computation time is wasted.

			Note the poor results that the generational genetic algorithm offers, which are due to the low diversity and fast convergence. It is known that this model can not reach the performance of CHC and Micro-GA with the classic reinitialization, see Table \ref{table:chc-mga}. Consequently, the performance's differences compared with those algorithms with greedy reinitialization are huge.

			GADEGD not only obtains high quality solutions but is also able to reach the problems' optimal solutions. We have developed Table \ref{table:state-of-the-art:convergence} in order to study how difficult is for the algorithms to converge to the instances' optimums.  Each entry contains the number of times that the corresponding algorithm has reached an optimal solution and the average time needed to do so. The results are taken from 30 executions per algorithm and instance, each of which lasts at most 20 seconds. GADEGD presents the fastest convergence. It also reaches the optimums more often than the other algorithms. The greedy diversification contributes to this convergence since it introduces new greedy chromosomes progressively, allowing the population's solutions to find the genetic material which they need to generate better descendants.

            \begin{table}[ht]
            	\caption{Convergence to the optimum solutions.}
            	\label{table:state-of-the-art:convergence}
            	\centering
            	\kern 2mm
            	\begin{tabular}{lllll}
            		\toprule
            		\textbf{Problem} & \multicolumn{4}{c}{\textbf{Heuristics}} \\
            		\midrule
            		\multicolumn{1}{c}{} & GADEGD & Classical GA & \descrcell{CHC}{with G.R.} & \descrcell{Micro-GA}{with G.R.} \\
            		\cmidrule{2-5}
            		berlin52 & 24 / \textbf{0.136} & Not reached & \textbf{28} / 0.35   & 23 / 0.67  \\
            		kroA100  & \textbf{11 / 3.95}  & Not reached & 8  / 12.576 & 6  / 10.77 \\
            		rd100    & \textbf{30 / 3.81}  & Not reached & 25 / 5.26   & 6  / 7.35  \\
            		\bottomrule
            	\end{tabular}
            \end{table}

\begin{figure}[H]
	        	\centering
	        	\includegraphics[width=10cm]{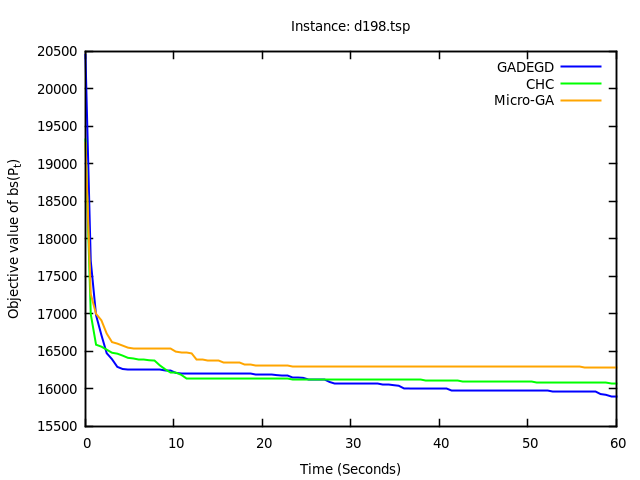}
	        	\caption{Convergence: GADEGD vs CHC with G.R. vs Micro-GA with G.R.}
	        	\label{fig:convergence}
	        \end{figure}

			Figure \ref{fig:convergence} shows how the algorithms' best solution evolve in the instance $d198$. The data is taken from a 60 seconds execution, plotting the objective value of the best solution found as time passes. The generational genetic algorithm is omitted in the study due to its bad performance. GADEGD, CHC and Micro-GA make a huge improvement to the initial solutions. However, after some iterations, they find more difficulties since the proportion of better chromosomes in the solutions space is getting smaller. At this point the exploitation of the best solutions' neighborhood and the capability to find new potential chromosomes are the most important qualities. The three algorithms have characteristics that help to achieve these purposes. However, Micro-GA's small population can be an encumbrance to fully achieve these qualities, being the algorithm with the worst performance. Furthermore, the reinitialization of both CHC and Mircro-GA makes the algorithm to start the search again, losing time in the process. GADEGD does not have these problems and it does actually find better solutions after 25 seconds, not falling into local optimums.

\begin{figure}[H]
        	\centering
        	\includegraphics[width=10cm]{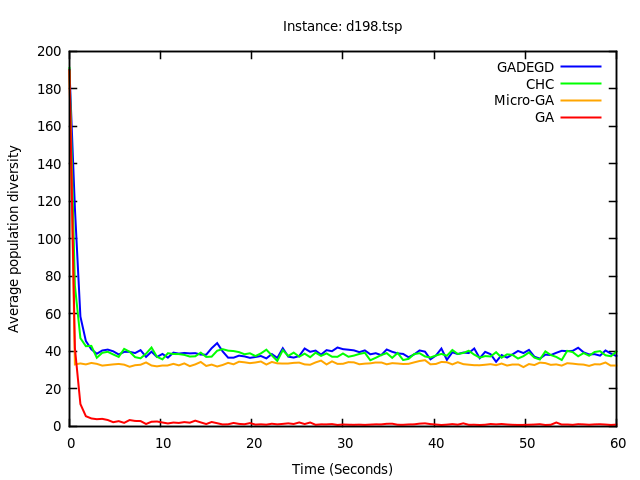}
        	\caption{Diversity: GADEGD vs GA (Algorithm \ref{alg:ga}) vs CHC with G.R. vs Micro-GA with G.R.}
        	\label{fig:diversity:all}
        \end{figure}

			Figure \ref{fig:diversity:all} shows how the population diversity evolves for the four algorithms studied: GADEGD, a generational GA and both CHC and Micro-GA with greedy reinitialization. The data corresponds to the execution given in Figure \ref{fig:convergence}. Each value is computed as the mean of the diversity in an interval of time. As we showed before, the generational genetic algorithm can not maintain a suitable diversity. On the other hand, GADEGD, CHC and Micro-GA present similar diversity in average thanks to the diversification and reinitialization operators. Note that the reinitialization procedure makes radical changes in the population and, as a consequence, the real diversity (not average) varies from zero to high values throughout the CHC and Micro-GA execution.

		\subsubsection{GADEGD's components analysis}

			One could wonder if GADEGD would perform equally well introducing random solutions in the diversification operator instead of greedy ones, what would increase the diversity even more. However, this model does not achieve the same results in practice. As we pointed out in Section \ref{sec:gadegd:gd}, greedy solutions contains a high quality genetic material that is transferred to its children and, after a few generations, spread to the whole population. The good performance of the hybridization between greedy and genetic algorithms is corroborated in Section \ref{sec:gadegd:experiments:state-of-the-art}, where we compared a reinitialization with greedy solutions with a randomized reinitialization for CHC and Micro-GA.

			Another important question is how the greedy diversification does actually influence the algorithm's performance. Table \ref{table:gadegd:function} showed that this mechanism generates between 2 and 5 per cent of the solutions for $g=Id$, what is an considerable amount of solutions. We introduce Table \ref{table:components:gd} in order to check if these solutions were important for the algorithm's results.

            \begin{table}[H]
            	\caption{Comparison among GADEGD, the same model without the greedy diversification, the genetic algorithm given in Algorithm\ref{alg:ga} and this model with greedy diversification. The execution time is $0.1 m$ seconds. The best results are highlighted in bold and the worst are underlined. The last row indicates the number of times that each model got the best and worst result in any instance.}
            	\label{table:components:gd}
            	\centering
            	\kern 2mm
            	\begin{tabular}{llllll}
            		\toprule
            		\textbf{Problem} & \textbf{Optimum} & \multicolumn{4}{c}{\textbf{Mean objective value}} \\
            		\midrule
            		\multicolumn{2}{c}{} & GADEGD & \descrcell{GADEGD}{without G.D.}    & GA & \descrcell{GA}{with G.D.} \\
            		\cmidrule{3-6}
            		eil51    & 426    & \textbf{427.267} & \underline{550.167} & 507.7 & 433.2 \\
            		berlin52 & 7542   & 7572.57 & \underline{9275.53} & 9146.5 & \textbf{7561.03} \\
            		st70     & 675    & \textbf{682.067} & \underline{1016.83} & 910.067 & 694.2 \\
            		eil76    & 538    & \textbf{549.5} & \underline{785.467} & 692.4 & 559.1 \\
            		pr76     & 108159 & \textbf{109395} & \underline{159413} & 146800 & 113510 \\
            		kroA100  & 21282  & \textbf{21352.5} & \underline{43765.6} & 34162.4 & 21747.8 \\
            		rd100    & 7910   & \textbf{7919.47} & \underline{15226.7} & 12421.3 & 8054.83 \\
            		eil101   & 629    & \textbf{633.3} & \underline{1082.43} & 886.833 & 657.167 \\
            		lin105   & 14379  & \textbf{14430.5} & \underline{28946.7} & 24862.4 & 14540.3 \\
            		ch150    & 6528   & \textbf{6578.67} & \underline{16518.7} & 12898.4 & 6676.47 \\
            		rat195   & 2323   & \textbf{2386.83} & \underline{6094.77} & 5246.8 & 2479 \\
            		d198     & 15780  & \textbf{16053.9} & \underline{36032.6} & 35969.5 & 16739.7 \\
            		ts225    & 126643 & \textbf{127427} & \underline{464102} & 394841 & 130436 \\
            		a280     & 2579   & \textbf{2704.5} & \underline{9365.33} & 8280.07 & 2847.17 \\
            		lin318   & 42029  & \textbf{43739.5} & \underline{177397} & 154852 & 46395 \\
            		fl417    & 11861  & \textbf{12303.9} & 81758.7 & \underline{97497.8} & 12741 \\
            		pcb442   & 50778  & \textbf{55502} & \underline{238401} & 233595 & 59163.1 \\
            		rat575   & 6773   & \textbf{7670.97} & 33968.8 & \underline{35167.7} & 7966.1 \\
            		\midrule
            		& & 17 / 0 & 0 / 16 & 0 / 2 & 1 / 0 \\
            		\cmidrule[0.3mm]{3-6}
            	\end{tabular}
            \end{table}

         	First, we have executed GADEGD without the greedy diversification operator. As one could expect, the algorithm's high pressure with no diversification scheme implies a very fast convergence and, thus, very poor results. Secondly, we have applied the greedy diversification to the generational genetic algorithm given in Algorithm \ref{alg:ga}. The results prove that the greedy diversification makes a huge positive impact in genetic algorithms' performance. However, as we indicated in Section \ref{sec:gadegd:gd}, the synergy among the components of this model was improvable in theory. The results also show that this synergy was increased in the GADEGD algorithm, which obtains the best results in 17 out of 18 instances.

			The competition between parent and children plays a crucial role in using the diversity efficiently since it allows to select and exploit the most promising region of the solutions' space. If an usual elitism is used instead of the competition scheme in GADEGD, then the diversity is not properly controlled and the algorithm results are not good enough as it is shown in Table \ref{table:components:selection-competition}. This table also includes the results obtained from a GADEGD version in which the binary tournament selection replaces the randomized adjacent selection. In this case the pressure applied to the population is excessive and the population diversity is partially lost, as we explained in Section \ref{sec:gadegd:gadegd}. Consequently, it can not reach the performance of GADEGD.

            \begin{table}[H]
            	\caption{Comparison with other pressure and selection mechanisms. The execution time is $0.1 m$ seconds. The best results are highlighted in bold and the worst are underlined. The last row indicates the number of times that each model got the best and worst result in any instance.}
            	\label{table:components:selection-competition}
            	\centering
            	\kern 2mm
				\begin{tabular}{lccc}
					\toprule
					\textbf{Problem} & \multicolumn{3}{c}{\textbf{Mean objective value}} \\
					\midrule
					& GADEGD & \descrcell{GADEGD}{without competition \\ between parent and children \\ and with elitism} & \descrcell{GADEGD}{with Tournament \\ selection} \\
					\midrule
					eil51    & \textbf{427.267} & \underline{452.067}  & 439.467 \\
					berlin52 & \textbf{7572.57} & 7770.83              & \underline{7902.8} \\
					st70     & \textbf{682.067} & \underline{730.333}  & 697.2 \\
					eil76    & \textbf{549.5}   & \underline{580.8}    & 569.4 \\
					pr76     & \textbf{109395}  & 119693               & \underline{120715} \\
					kroA100  & \textbf{21352.5} & \underline{23218.6}  & 22749.8 \\
					rd100    & \textbf{7919.47} & \underline{8386.87}  & 8324.43 \\
					eil101   & \textbf{633.3}   & \underline{678.9}    & 648.233 \\
					lin105   & \textbf{14430.5} & \underline{14835}    & 14633.2 \\
					ch150    & \textbf{6578.67} & 6876.27              & \underline{6940.2} \\
					rat195   & \textbf{2386.83} & 2523                 & \underline{2534.37} \\
					d198     & \textbf{16053.9} & \underline{17362.9}  & 16980.9 \\
					ts225    & \textbf{127427}  & 133066               & \underline{134465} \\
					a280     & \textbf{2704.5}  & \underline{2896}     & 2852.07 \\
					lin318   & \textbf{43739.5} & 48052.4              & \underline{48773} \\
					fl417    & \textbf{12303.9} & \underline{13451.9}  & 12852.8 \\
					pcb442   & \textbf{55502}   & 59581.4              & \underline{60669.3} \\
					rat575   & \textbf{7670.97} & 8027.63              & \underline{8057.63} \\
					\midrule
					& 18 / 0 & 0 / 10 & 0 / 8 \\
					\cmidrule[0.3mm]{1-4}
				\end{tabular}
            \end{table}

            In summary, each GADEGD's component is relevant for the algorithm's performance. The cooperation among all the introduced components allows to achieve a healthy diversity and an equilibrium between exploration and exploitation.

\section{Memetic algorithm with diversity equilibrium based on greedy diversification} \label{sec:madegd}

	In this section we extend GADEGD to the field of memetic algorithms. First, we argue how to define this new metaheuristic, called MADEGD. Secondly, we develop an experimental study in which MADEGD's behaviour is analysed and compared with other state of the art heuristics based on local search.

	\subsection{Memetic algorithm with diversity equilibrium based on greedy diversification} \label{sec:madegd:madegd}

		MADEGD is obtained when GADEGD is hybridized with a local search procedure, as it is done in memetic algorithms. In Section \ref{sec:ga-ma:ma} we argued that a good hybridization is applying the local search once per iteration to the best population's chromosome that has not been improved before. Hence, we use this scheme in MADEGD. However, we must decide whether the greedy diversification operator is applied before of after the local search. We choose to use the greedy diversification first in order to avoid that a repeated solution introduced by a crossover is improved.

		Another important question is how to initialize the population. If the population were randomly chosen, then the local search would be applied to very low quality solution in the initial iterations, what consumes too much time. Therefore, we initialize the population with solutions obtained by a greedy randomized algorithm as we did in Algorithm \ref{alg:ma}.

		Lastly, GADEGD has two parameters, the characteristic function and the population size. GADEGD obtained the best results when the characteristic function was $g = Id$. Hence, we use this function in MADEGD. The population size is analyzed in Section \ref{sec:madegd:ea:ps}.

		\begin{algorithm}
			\begin{algorithmic}[1]
				\scriptsize
				\caption{Memetic algorithm with diversity equilibrium based on greedy diversification} \label{alg:madegd}
				\REQUIRE The population size, named $n$.
				\STATE Initialize $P_0$ with $n$ solutions obtained by a greedy randomized algorithm.
				\STATE $t \leftarrow 0$
				\WHILE {stopping criteria is not achieved}
				\STATE $P_{t+1} \leftarrow BuildNewPopulationGADEGD(P_t)$
				\STATE $P_{t+1} \leftarrow GreedyDiversification(P_{t+1}, Id)$
				\STATE Apply the local search to the best solution not previously improved of P t+1 (if it exits).
				\STATE $t \leftarrow t+1$
				\ENDWHILE
				\RETURN $bs(P_t)$
			\end{algorithmic}
		\end{algorithm}

		Algorithm \ref{alg:madegd} shows the pseudo-code of MADEGD. Note that if a greedy solution is added to MADEGD's population, then it will be crossed with the population's solutions (which are presumably better) until it is good enough to be improved by the local search. Consequently, the algorithm is finding potential chromosomes which are in the path between various greedy and high quality population's solutions. This fact will allow the local search to perform the best it is able to.

		The application of MADEGD to the traveling salesman problem is straightforward. The greedy randomized algorithm is the same used in the greedy diversification (see Section \ref{sec:gadegd:gd}). Furthermore, we use Lin-Kernighan as the local search procedure.

	\subsection{Experimental analysis} \label{sec:madegd:ea}

		The experimental analysis contains 3 subsections. First, we study how the population size affects MADEGD. Secondly, we compare it with GADEGD in order to understand how the local search change the algorithm's behaviour, contrasting its better performance. Thirdly, MADEGD is matched against another memetic algorithm, GRASP and iterated greedy from a triple perspective, solutions quality, population diversity and calls to the local search.

		\subsubsection{Analysis of the population size} \label{sec:madegd:ea:ps}

			In Section \ref{sec:gadegd:experiments:parameters} we mentioned how important the population size is for a genetic algorithm. The same arguments are valid in the field of memetic algorithms. Table \ref{table:madegd:ps} contains the results obtained by MADEGD with population sizes 8, 16, 32 and 64. Note that the performance is better when the population is smaller. The reason is that most of the computational time is wasted in the local search. Consequently, less iterations of the genetic operators are applied and a higher pressure is needed, what is provided by the smaller population size.

            \begin{table}[H]
				\caption{MADEGD with different population sizes. The execution time is $0.1 m$ seconds. The best results are highlighted in bold and the worst are underlined. The last row indicates the number of times that each model got the best and worst result in any instance.}
				\label{table:madegd:ps}
				\centering
				\kern 2mm
				\begin{tabular}{llllll}
					\toprule
					\textbf{Problem} & \textbf{Optimum} & \multicolumn{4}{c}{\textbf{Mean objective value}} \\
					\midrule
					& & $n = 8$ & $n = 16$ & $n = 32$ & $n = 64$ \\
					\cmidrule{3-6}
					eil51    & 426    & \textbf{426} & 426.167 & 426.867 & \underline{426.933} \\
					berlin52 & 7542   & \textbf{7542} & \textbf{7542} & \textbf{7542} & \textbf{7542} \\
					st70     & 675    & \textbf{675} & \textbf{675} & 676 & \underline{677.767} \\
					eil76    & 538    & \textbf{538} & \textbf{538} & 538.133 & \underline{538.333} \\
					pr76     & 108159 & \textbf{108159} & \textbf{108159} & \textbf{108159} & \textbf{108159} \\
					kroA100  & 21282  & \textbf{21282} & \textbf{21282} & \textbf{21282} & \underline{21282.8} \\
					rd100    & 7910   & \textbf{7910} & \textbf{7910} & \textbf{7910} & \underline{7916.53} \\
					eil101   & 629    & \textbf{629} & \textbf{629} & 629.533 & \underline{630.267} \\
					lin105   & 14379  & \textbf{14379} & \textbf{14379} & \textbf{14379} & \textbf{14379} \\
					ch150    & 6528   & \textbf{6529.73} & 6541.5 & 6547.3 & \underline{6549.63} \\
					rat195   & 2323   & \textbf{2326.67} & 2329.4 & 2331.83 & \underline{2334.47} \\
					d198     & 15780  & \textbf{15794.7} & 15801.4 & 15805.5 & \underline{15815.3} \\
					ts225    & 126643 & \underline{127036} & 126794 & \textbf{126791} & 126895 \\
					a280     & 2579   & \textbf{2582.8} & \textbf{2582.8} & 2590.8 & \underline{2596.33} \\
					lin318   & 42029  & 42349.4 & \textbf{42300} & 42376.4 & \underline{42444.9} \\
					fl417    & 11861  & \underline{11948.8} & \textbf{11940.8} & 11948.7 & 11945.6 \\
					pcb442   & 50778  & 51438.2 & \textbf{51257.1} & 51438.3 & \underline{51593.8} \\
					rat575   & 6773   & 6878.73 & 6874.23 & \textbf{6869.27} & \underline{6878.77} \\
					\midrule
					& & 13 / 2 & 12 / 0 & 8 / 0 & 3 / 13 \\
					\cmidrule[0.3mm]{3-6}
				\end{tabular}
			\end{table}

            We have executed the algorithm with greater execution times concluding that the size 8 is not good enough because it does not provide sufficient exploration of the solutions space. As a consequence, we propose 16 as the standard population size for MADEGD.

            Note that population based heuristics usually need a bigger population to avoid premature convergence. However, MADEGD does not necessary needs a big size thanks to the greedy diversification.

		\subsubsection{Comparison with the genetic algorithm with diversity equilibrium based on greedy diversification} \label{sec:madegd:gadegd}

			Table \ref{table:madegd:gadegd} presents the results obtained by MADEGD and GADEGD with population sizes 16 and 64 respectively. The mean objective value of the solutions obtained by MADEGD is drastically better, what shows how effective is the local search combined with the genetic operators and the greedy diversification. In fact, MADEGD finds the optimum solutions in most instances.

			In Section \ref{sec:gadegd:experiments:parameters} it was noticed that GADEGD needs 64 as the population size to keep an equilibrium between exploration and exploitation. However, as it is pointed out previously, MADEGD requires a smaller population because it performs less iterations. This statement is corroborated in Table \ref{table:madegd:gadegd}, which provides the average number of solutions computed in each instance by both algorithms. GADEGD generates between 10 and 50 more solutions than MADEGD. However, MADEGD iterations are much more effective thanks to the local search.

	        \begin{table}[H]
	        	\caption{MADEGD vs GADEGD with population sizes 16 y 64 respectively. The execution time is $0.1 m$ seconds. The best results are highlighted in bold and the worst are underlined. The last row indicates the number of times that each model got the best result in any instance.}
	        	\label{table:madegd:gadegd}
	        	\centering
	        	\kern 1mm
	        	\begin{tabular}{llllll}
	        		\toprule
	        		\textbf{Problem} & \textbf{Optimum} & \multicolumn{2}{c}{\textbf{Mean objective value}} &   \multicolumn{2}{c}{\textbf{Number of generated}} \\
		        	& & \multicolumn{2}{c}{} &   \multicolumn{2}{c}{\textbf{solutions}} \\
	        		\midrule
	        		& & MADEGD & GADEGD & MADEGD & GADEGD \\
	        		\cmidrule{3-4} \cmidrule{5-6}
	        		eil51    & 426    & \textbf{426.167} & \underline{427.267} & 128759  & 1692820 \\
	        		berlin52 & 7542   & \textbf{7542}    & \underline{7572.57} & 46387.8 & 1731320 \\
	        		st70     & 675    & \textbf{675}     & \underline{682.067} & 114622  & 1674870 \\
	        		eil76    & 538    & \textbf{538}     & \underline{549.5}   & 127619  & 1740730 \\
	        		pr76     & 108159 & \textbf{108159}  & \underline{109395}  & 60987.1 & 1377370 \\
	        		kroA100  & 21282  & \textbf{21282}   & \underline{21352.5} & 51970.3 & 14260   \\
	        		rd100    & 7910   & \textbf{7910}    & \underline{7919.47} & 54622.7 & 1473510 \\
	        		eil101   & 629    & \textbf{629}     & \underline{633.3}   & 92876.7 & 1407060 \\
	        		lin105   & 14379  & \textbf{14379}   & \underline{14430.5} & 34887.5 & 538391  \\
	        		ch150    & 6528   & \textbf{6541.5}  & \underline{6578.67} & 50950   & 1270930 \\
	        		rat195   & 2323   & \textbf{2329.4}  & \underline{2386.83} & 47671   & 379744  \\
	        		d198     & 15780  & \textbf{15801.4} & \underline{16053.9} & 16071.4 & 362111  \\
	        		ts225    & 126643 & \textbf{126794}  & \underline{127427}  & 53980.6 & 724328  \\
	        		a280     & 2579   & \textbf{2582.8}  & \underline{2704.5}  & 45372.8 & 612916  \\
	        		lin318   & 42029  & \textbf{42300}   & \underline{43739.5} & 10964.5 & 485157  \\
	        		fl417    & 11861  & \textbf{11940.8} & \underline{12303.9} & 5499.87 & 422658  \\
	        		pcb442   & 50778  & \textbf{51257.1} & \underline{55502}   & 16309   & 231215  \\
	        		rat575   & 6773   & \textbf{6874.23} & \underline{7670.97} & 3522.03 & 125132  \\
					\midrule
					& & 18  & 0 &  &  \\
					\cmidrule[0.3mm]{3-6}
	        	\end{tabular}
	        \end{table}

		\subsubsection{Comparison with other local search based multi-start metaheuristics}

	        Local search based multi-start metaheuristics try to apply the local search to promising solutions placed in different regions of the search space. Consequently, they require an underneath procedure which supplies high quality and diverse solutions on which local search will be executed. Hence, local search based metaheuristics can be understood as an hybridization between local search and other heuristics.

	        Memetic algorithms are local search based multi-start metaheuristics in which the local search is applied to new solutions obtained by the evolutionary operators. This hybridization presents several advantages. First, the evolution scheme guarantees that the local search will be applied to better solutions as time passes. Secondly, the solutions obtained by the local search contain some information which can be used in the evolutionary algorithm's iterations, obtaining a better performance. However, if the evolutionary scheme doesn't pay enough attention to the exploration, then the local search is applied to similar solutions over and over. Consequently, it might always find the same local optimums and the computation time is wasted. Hence, the evolutionary scheme should have a good equilibrium between exploration and exploitation in order to obtain a high performance memetic algorithm.

	        Other local search based multi-start metaheuristics, such as GRASP \cite{grasp} and iterated greedy \cite{iterated-greedy}, \cite{ig-tsp}, use techniques founded on randomized greedy algorithms. Greedy solutions are placed in promising regions of the solutions space and, thus, the local search is highly productive on them.

	        Algorithm \ref{alg:grasp} describes how GRASP works, at each iteration a greedy solution is obtained by a randomized greedy algorithm and it is improved by the local search. The best solution found is returned at the end of the algorithm.

			\begin{algorithm}
				\begin{algorithmic}[1]
					\scriptsize
					\caption{GRASP} \label{alg:grasp}
					\WHILE {stopping criteria is not achieved}
					\STATE $s \leftarrow GreedyRandomizedAlgorithm()$
					\STATE $s' \leftarrow LocalSearch(s)$
					\IF{$s'$ is the best solution found}
					\STATE $best\_solution \leftarrow s'$
					\ENDIF
					\ENDWHILE
					\RETURN $best\_solution$
				\end{algorithmic}
			\end{algorithm}

	         GRASP does not use the information obtained in the past computations, its iterations are independent and equally productive in average. Iterated greedy try to overcome this issue modifying a previously visited solution with a greedy technique in order to create new elements of the solution space. Algorithm \ref{alg:ig} provides an usual implementation of iterated greedy. At each step, a destruction procedure is applied to the best solution found. The destruction procedure removes a subset of the solution's data, obtaining a partial solution. Afterwards, the partial solution is reconstructed by a randomized greedy technique and the obtained solution is improved by the local search.

			\begin{algorithm}
				\begin{algorithmic}[1]
					\scriptsize
					\caption{Iterated Greedy} \label{alg:ig}
					\STATE $best\_solution \leftarrow GreedyRandomizedAlgorithm()$
					\STATE $best\_solution \leftarrow LocalSearch(best\_solution)$
					\WHILE {stopping criteria is not achieved}
					\STATE $s \leftarrow Destruction(best\_solution)$
					\STATE $s' \leftarrow RandomizedGreedyReconstruction(s)$
					\STATE $s'' \leftarrow LocalSearch(s')$
					\IF{$s''$ is the best solution found}
					\STATE $best\_solution \leftarrow s''$
					\ENDIF
					\ENDWHILE
					\RETURN $best\_solution$
				\end{algorithmic}
			\end{algorithm}

	        Particularizing in the traveling salesman problem, the destruction procedure consist in removing a random sublist of the solution's representation. The reconstruction step is carried out by the randomized greedy algorithm based on the nearest neighbor philosophy which was introduced in Section \ref{sec:gadegd:gd}. We have implemented GRASP and iterated greedy using Lin-Kernighan as the local search procedure.

			Note that the destruction - reconstruction step of iterated greedy can be understood as a crossover between the best solution found and a greedy solution. Thus, we can see iterated greedy as an hybridization between greedy and memetic algorithms. From this perspective the model is improvable in terms of exploration and exploitation. The population size is 1 and, thus, it usually operates in the same region of the search space. Furthermore, the local search is always applied in each iteration even if the obtained solution is not good enough. Hence, we can conclude that iterated greedy is more focused on exploitation than on exploration.

            \begin{table}[H]
       			\caption{Comparison of MADEGD, MA (Algorithm \ref{alg:ma}), GRASP and IG. The execution time is $0.1 m$ seconds. The best results are highlighted in bold and the worst are underlined. The last row indicates the number of times that each model got the best and worst result in any instance.}
       			\label{table:ls-algorithms}
       			\centering
       			\kern 1mm
       			\begin{tabular}{llllll}
       				\toprule
       				\textbf{Problem} & \textbf{Optimum} & \multicolumn{4}{c}{\textbf{Mean objective value}} \\
       				\midrule
       				& & MADEGD & MA & GRASP & IG \\
       				\cmidrule{3-6}
       				eil51    & 426    & \textbf{426.167} & \underline{433.8} & 432.133 & 432.1 \\
       				berlin52 & 7542   & \textbf{7542} & 7628.8 & \underline{7670.43} & 7665.73 \\
       				st70     & 675    & \textbf{675} & 689.567 & \underline{692.433} & 692.133 \\
       				eil76    & 538    & \textbf{538} & 548 & \underline{551.867} & 550.733 \\
       				pr76     & 108159 & \textbf{108159} & 109382 & 110083 & \underline{110544} \\
       				kroA100  & 21282  & \textbf{21282} & 21408 & 21469.2 & \underline{21475.8} \\
       				rd100    & 7910   & \textbf{7910} & 7951.87 & 8033.37 & \underline{8061.1} \\
       				eil101   & 629    & \textbf{629} & 641.533 & 647.833 & \underline{648.267} \\
       				lin105   & 14379  & \textbf{14379} & 14482.7 & 14551.6 & \underline{14569.7} \\
       				ch150    & 6528   & \textbf{6541.5} & 6575.73 & 6642.1 & \underline{6671.73} \\
       				rat195   & 2323   & \textbf{2329.4} & 2349.37 & \underline{2373.23} & 2369.23 \\
       				d198     & 15780  & \textbf{15801.4} & 15883.3 & \underline{16015} & 15981.3 \\
       				ts225    & 126643 & \textbf{126794} & 129022 & 129865 & \underline{130035} \\
       				a280     & 2579   & \textbf{2582.8} & 2637.43 & 2659.3 & \underline{2660.7} \\
       				lin318   & 42029  & \textbf{42300} & 42827.2 & 43068.8 & \underline{43157.7} \\
       				fl417    & 11861  & \textbf{11940.8} & 12041.1 & \underline{12156.8} & 12130.5 \\
       				pcb442   & 50778  & \textbf{51257.1} & 52319.9 & \underline{52589.2} & 52585 \\
       				rat575   & 6773   & \textbf{6874.23} & 6924.53 & 6944.6 & \underline{6951.93} \\
       				\midrule
       				& & 18 / 0 & 0 / 1 & 0 / 7 & 0 / 10 \\
       				\cmidrule[0.3mm]{3-6}
       			\end{tabular}
       		\end{table}

            MADEGD is also an hybridization between greedy and memetic algorithms. However, it combines the best of both worlds. The greedy diversification promotes the exploration of the promising regions of the search space when it is needed. Furthermore, the competition between parent and children control the population's quality and, as a consequence, the local search is likely applied to better solutions each iteration. As we mentioned before, if a greedy solution enters in the population, then it will get crossed with better chromosomes until it is good enough to be improved by the local search.

	        This synergy is the reason behind the results presented in Table \ref{table:ls-algorithms}, which compares the performance of MADEGD, the memetic algorithm given in Algorithm \ref{alg:ma} (MA), GRASP and iterated greedy (IG). MADEGD loosely obtains the best result in every instance. MA also outperforms GRASP and IG thanks to the evolutionary character. Note that GADEGD's results are better than the MA's ones in instances with less than 110 cities (see Table \ref{table:madegd:gadegd}) in spite of implementing a version of Lin-Kernighan as the local search, one of the best heuristics for the travelling salesman problem.

       		Table \ref{table:ls-calls} provides the average number of calls to the local search in Table \ref{table:ls-algorithms}'s executions. MA is the heuristic which presents more calls to the local search. This fact is due to the population convergence, the local search is much faster because the solutions to which it is applied are near to local optimums. GRASP and IG are the algorithms with less number of calls to the local search. Each iteration of those algorithms consists in applying the local search to a greedy or partially greedy solution, what is very time consuming since there is a lot of room for optimization. MADEGD mixes the best of both worlds again since it constantly explores the search space but the local search is only applied to the best possible solutions.

            \begin{figure}[H]
       			\centering
       			\includegraphics[width=10cm]{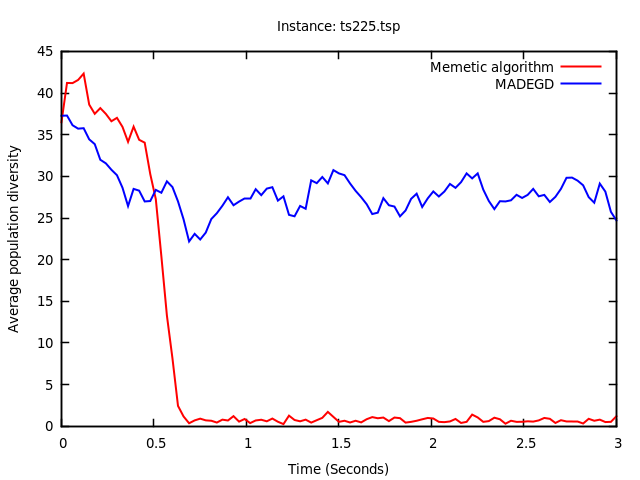}
       			\caption{Diversity: MADEGD vs Memetic algorithm (Algorithm \ref{alg:ma}).}
       			\label{fig:madegd:diversity}
       		\end{figure}

            The number of calls to the local search of GRASP and IG is equal to the number of generated solutions. However, both MA and MADEGD just apply the local search in an iteration if there is a solution no previously improved by the local search. Table \ref{table:ls-calls} also shows the percent of iterations in which the local search was applied to a population's solution for both memetic algorithms. MA always applies the local search since the population is almost fully generated by the crossover operator. However, MADEGD, after a fair number of iterations, only applies the local search if a new solution has entered the population after the competition between parent and children. Nonetheless, the percent is always greater than 85\%, what shows that the crossover operator is able to find better chromosomes than the parents. This fact is essential for the algorithm's good behavior since if the crossover were not good enough, then no solution would enter the population and the algorithm would converge to a local optimum.

            Finally, Figure \ref{fig:madegd:diversity} shows how the diversity evolves in an execution of MADEGD and the memetic algorithm. It is very similar to Figure \ref{fig:diversity:gadegd}. The memetic algorithm converges too fast to a local optimum, what avoids a proper exploration of the search space.

            \begin{table}[H]
       			\caption{Average number of calls to the local search in the executions of Table \ref{table:ls-algorithms}. Percent of MADEGD and memetic algorithm (MA)'s iterations in which the local search is applied to a population's solution.}
       			\label{table:ls-calls}
       			\centering
       			\kern 1mm
       			\begin{tabular}{lllllllc}
       				\toprule
       				\textbf{Problem} & \multicolumn{4}{c}{\textbf{Number of calls}} & \multicolumn{3}{c}{\textbf{Percent of iterations}} \\
       				& \multicolumn{4}{c}{\textbf{to the local search}} & \multicolumn{3}{c}{\textbf{in which the local}} \\
       				& \multicolumn{4}{c}{} & \multicolumn{3}{c}{\textbf{search is applied}} \\
       				\midrule
       				& MADEGD & MA & GRASP & IG & & MADEGD & MA \\
       				\cmidrule{2-5} \cmidrule{7-8}
       				eil51    & 7709.63 & 7478.47 & 3309.6 & 3256.13  & & 100     & 100    \\
       				berlin52 & 2759 & 4054.87 & 1863.37 & 1825.3     & & 100     & 100    \\
       				st70     & 6869.37 & 6690.73 & 2269.07 & 2247.83 & & 100     & 100    \\
       				eil76    & 7682.43 & 7756.63 & 2514.67 & 2479.27 & & 100     & 100    \\
       				pr76     & 3653.83 & 4940.3 & 722.633 & 712.6    & & 100     & 100    \\
       				kroA100  & 3096.7 & 4161.97 & 942.133 & 931.267  & & 100     & 100    \\
       				rd100    & 3268.77 & 4268.37 & 854.7 & 843.933   & & 100     & 100    \\
       				eil101   & 5543.07 & 5605.2 & 1390.47 & 1363.07  & & 100     & 100    \\
       				lin105   & 2085.3 & 3203.63 & 392.1 & 394.8      & & 100     & 100    \\
       				ch150    & 3067.7 & 3762.7 & 765.433 & 760       & & 100     & 100    \\
       				rat195   & 2871.6 & 4212.27 & 394.133 & 393.533  & & 100     & 100    \\
       				d198     & 968.467 & 1381.4 & 119.9 & 119.567    & & 100     & 100    \\
       				ts225    & 3284.07 & 4003.83 & 746.9 & 736.4     & & 99.9918 & 100    \\
       				a280     & 2729.97 & 3423.83 & 240.3 & 239.733   & & 100     & 100    \\
       				lin318   & 663.267 & 1093.1 & 56.4 & 56.0333     & & 99.7588 & 100    \\
       				fl417    & 331.533 & 586.8 & 33.0667 & 33.4333   & & 100     & 100    \\
       				pcb442   & 992.133 & 1765.13 & 74.1 & 74.9333    & & 97.6112 & 100    \\
       				rat575   & 216.3 & 1059.77 & 28.9667 & 27.8333   & & 85.2672 & 100    \\
       				\bottomrule
       			\end{tabular}
       		\end{table}

\section{Conclusion} \label{sec:conclusion}

	In this paper we have introduced a novel genetic algorithm, GADEGD, which attempts to achieve a balance between exploration and exploitation. The algorithm's key operator is the greedy diversification, which maintains a diversity equilibrium in the population. Furthermore, the algorithm uses the randomized adjacent selection and a competition between parent and children.  These operators have been selected in order to increase the components' synergy.

	We have also extended the algorithm to the field of memetic algorithms, MADEGD, obtaining a more competitive metaheuristic which outperforms a generational memetic algorithm, GRASP and iterated greedy in our studies.

	The greedy diversification has been proved to be a relevant operator for designing population based metaheuristics and, in particular, genetic and memetic algorithms. An heuristic which uses this operator has much more facilities to constantly keep a high quality and diverse population, what can not be achieved by the widely used mutation operator.

	The developed work reaffirms our initial assertions, the equilibrium between exploration and exploitation and the diversity problem should be taken into account when designing genetic and memetic algorithms. Hybridization helps to solve both problems, providing exploration and exploitation mechanisms to evolutionary algorithms.

	Finally, we believe that the proposed metaheuristics and operators can be fruitfully applied to high dimensional or large scale problems \cite{ma:high-dimensions}, \cite{mh:large-scale}, where memetic algorithms are one of the most powerful metaheuristics. These problem require a careful exploration of the search space and an effective exploitation of the best solutions found. Therefore, as a future work we will be extending the current results to the large scale framework.

\printbibliography

\end{document}